\documentclass[twocolumn]{svjour3}

\usepackage{amssymb,amsmath,amsfonts,graphicx}
\usepackage[english]{babel}
\usepackage{theorem}
\usepackage{algorithm,algorithmic}

\def\Div{\textup{div\,}}

\title{Bregman implementation of Meyer's $G-$norm for cartoon + textures decomposition}
\author{J\'er\^ome Gilles and Stanley Osher}
\institute{University of California Los Angeles, Department of Mathematics, 520 Portola Plaza,Los Angeles, CA 90095-1555\\ \email{jegilles@math.ucla.edu,sjo@math.ucla.edu}}

\date{}

\begin{document}
\maketitle

\begin{abstract}
In this paper, we design a very simple algorithm based on Split Bregman iterations to numerically solve the cartoon + textures decomposition model of Meyer. This results in a significant gain in speed compared to Chambolle's nonlinear projectors.
\keywords{Split Bregman, Meyer $G-$norm, Total variation, image decomposition}
\end{abstract}

\section{Introduction}
In the last few years, cartoon + textures image decomposition models have received a lot of interest from the image processing community. The original model was theoretically proposed by Yves Meyer in \cite{meyer} and used the total variation (TV) to model the cartoon part and a specific space $G$ for oscillatory patterns. Many developments, especially the work of Aujol et al. \cite{aujol-chambolle} who proposed an efficient way to numerically solve this model, appeared in the literature. The significant idea involves using a duality argument and the nonlinear projectors proposed by Chambolle \cite{chambolle} to find the oscillatory component which lies in the space $G$ defined by Meyer.\\
Recent works of Goldstein et al. \cite{Goldstein2009} in optimization show that Bregman iteration is a very efficient and fast way to solve TV problems. Surprisingly, in our knowledge, no work was proposed in the literature to design an algorithm which solves $G-$norm based problems by the use of Bregman iterations (and the same duality argument used by Chambolle).\\
In this paper, we describe the algorithm which aims to find the minimizer of a functional based on Meyer's $G-$norm by using Split Bregman iterations. The remainder of the paper is as follows: in section \ref{sec1}, we set up the basic notations used throughout the paper. We recall the definition of Chambolle's nonlinear projector and the cartoon + textures decomposition model. In section \ref{sec:BTV}, we give the algorithm designed by Goldstein et al. to solve TV problems by Split Bregman iterations. In section \ref{sec:BregG}, we describe how to use the Split Bregman iteration to solve $G-$norm models and then give the corresponding algorithm to do the cartoon + textures decomposition. Section \ref{sec:expe} presents some results obtained by the new algorithm. We conclude this work in section \ref{sec:conc}.

\section{Chambolle's projector and cartoon + textures decomposition}\label{sec1}
In \cite{chambolle}, A.Chambolle proposed a nonlinear projector to efficiently solve the Rudin-Osher-Fatemi (ROF) model \cite{rof} given in Eq.~\ref{eq:rof}.
\begin{equation}\label{eq:rof}
\hat{u}=\arg_{u}\min J(u)+\frac{\lambda}{2}\|f-u\|_2^2
\end{equation}
where $f$ is the original image, $J(u)$ is the total variation (TV) of $u$ (defined as $J(u)=\int|\nabla u|$) and $\hat{u}$ the restored image. This algorithm is widely used for denoising and restoration purposes because of its ability to preserve piecewise smooth functions. Total variation is directly related to the space of bounded variations ($BV$). If we consider the closure of $BV$ in $\mathcal{S}^2$ (denoted $\mathcal{BV}$), we can prove the existence of a dual space $G$ (sometimes denoted $\mathcal{BV}^*$). Considering the set $G_{\mu}=\left\{g\in G/ \|g\|_G\leqslant\mu\right\}$, Chambolle shows that the solution of the ROF model can be written as $\hat{u}=f-P_{G_{1/\lambda}}(f)$ where $P_{G_{\mu}}(f)$ corresponds to the projection onto $G_{\mu}$ . Practically, this nonlinear projector is evaluated by the help of Proposition.~\ref{prop:pg} (see \cite{chambolle} for details and proof).

\begin{proposition}\label{prop:pg}
If $\tau<\frac{1}{8}$ then $\mu \Div(p^n)$ converges to $P_{G_{\mu}}(f)$ when $n\rightarrow +\infty$ where
\begin{equation}\label{eq:p}
p_{i,j}^{n+1}=\frac{p_{i,j}^n+\tau \left(\nabla \left(\Div(p^n)-\frac{f}{\mu}\right)\right)_{i,j}}{1+\tau \left|\left(\nabla \left(\Div(p^n)-\frac{f}{\mu}\right)\right)_{i,j}\right|}
\end{equation}
\end{proposition}
In \cite{aujol-chambolle}, the authors proposed using this projector to do the cartoon + texture decomposition of an image.  They followed the idea of Meyer \cite{meyer} who showed that the space $G$ is well adapted to capture oscillatory patterns and then proposed to replace the $L^2$ norm in the ROF model by the $G-$norm. The corresponding cartoon ($u$) + texture ($v$) decomposition model is equivalent to minimize the problem given in Eq.~\ref{eq:tvg} for $(u,v)\in BV\times G_{\mu}$.

\begin{equation}\label{eq:tvg}
F_{\lambda,\mu}^{AU}(u,v)=J(u)+J^*\left(\frac{v}{\mu}\right)+\frac{\lambda}{2}\|f-u-v\|_{L^2}^2
\end{equation}

The authors of \cite{aujol-chambolle} then showed that the solution of this minimization problem is given by Algorithm.~\ref{alg:uv}.

\begin{algorithm}[!t]
\caption{Cartoon + textures decomposition algorithm based on Chambolle's projectors.}
\label{alg:uv}
\begin{algorithmic}
\STATE Initialisation: $u_0=v_0=0$
\WHILE {`` Not converged''}
\STATE Update $v$ by $v_{n+1}=P_{G_{\mu}}(f-u_n)$
\STATE Update $u$ by $u_{n+1}=f-v_{n+1}-P_{G_{1/\lambda}}(f-v_{n+1})$
\ENDWHILE
\end{algorithmic}
\end{algorithm}

This method works well to separate textures from cartoon and was adapted in different cases like in the presence of a convolution kernel or made locally adaptive \cite{Aujol2006b,Gilles2007e}.

\section{Split Bregman iteration for $TV$ minimization}\label{sec:BTV}
Recently a new minimization approach, called the Split Bregman iteration, was proposed by Goldstein et al. \cite{Goldstein2009} and is particularly well-adapted for $L^1$ schemes like the total variation. It is easy to implement and converges quickly (very few iterations are needed). If we denote $d_x=\nabla_xu$ and $d_y=\nabla_yu$ the derivatives with respect to $x$ and $y$ respectively, the problem described by Eq.~\ref{eq:rof} can be rewritten as the one depicted in Eq.~\ref{eq:breg}.

\begin{align}\label{eq:breg}
(\hat{d_x},\hat{d_y},\hat{u})&=\arg\min \sqrt{|d_x|^2+|d_y|^2}+\frac{\lambda}{2}\|u-f\|_2^2\\ \notag
&+\frac{\eta}{2}\|d_x-\nabla_xu-b_x\|_2^2+\frac{\eta}{2}\|d_y-\nabla_yu-b_y\|_2^2
\end{align}

The authors of \cite{Goldstein2009}, show that $(\hat{d_x},\hat{d_y},\hat{u})$ can be computed by the Algorithm.~\ref{algo:IROF}.

\begin{algorithm}
\caption{ROF restoration by Split Bregman iterations.}
\label{algo:IROF}
\begin{algorithmic}
\STATE $u^0=f,d_x^0=0,d_y^0=0,b_x^0=0,b_y^0=0$
\WHILE {``Not converged''}
\STATE Update $u^{k+1}$ by using equation (\ref{eq:rofustep})
\STATE Compute $s^k=\sqrt{|\nabla_xu^k+b_x^k|^2+|\nabla_yu^k+b_y^k|^2}$
\STATE $d_x^{k+1}=max(s^k-1/\eta),0)\frac{\nabla_xu^k+b_x^k}{s^k}$
\STATE $d_y^{k+1}=max(s^k-1/\eta),0)\frac{\nabla_yu^k+b_y^k}{s^k}$
\STATE $b_x^{k+1}=b_x^k+\nabla_xu^{k+1}-d_x^{k+1}$
\STATE $b_y^{k+1}=b_y^k+\nabla_yu^{k+1}-d_y^{k+1}$
\ENDWHILE
\end{algorithmic}
\end{algorithm}

where $u$ can be updated in the Fourier domain by Eq.~\ref{eq:rofustep} (in \cite{Goldstein2009}, Goldstein et al. propose to use a discrete version of the Laplacian and a Gauss-Seidel scheme instead of the Fourier domain), where the hat symbol stands for the Fourier transform and $\Re(g)$ the real part of $g$:

\begin{equation}\label{eq:rofustep}
\hat{U}=(\lambda\hat{I}-\eta\Re(\hat{\Delta}))^{-1}\left[\lambda\hat{F}-\eta\left(\Div(d^k-b^k)\right)^{\wedge}\right]
\end{equation}

\section{Bregman Meyer's $G-$norm implementation}\label{sec:BregG}
In this section, we show that finding the function $v\in G_{\mu}$ which minimize the model presented in Eq.~\ref{eq:gnorm}.

\begin{equation}\label{eq:gnorm}
J^*\left(\frac{v}{\mu}\right)+\frac{\lambda}{2}\|f-v\|_2^2
\end{equation}

can be done by using a duality argument and the Split Bregman iteration (in fact, we follow the proof of Chambolle but in the oppposite direction). Proposition~\ref{prop:g} gives the corresponding result.

\begin{proposition}\label{prop:g}
The function $v\in G_{\mu}$ which minimizes Eq.~\ref{eq:gnorm} is given by
\begin{equation}
\hat{v}=f-\frac{1}{\lambda}P_{ROF}\left(\lambda f,\frac{1}{\lambda\mu}\right)
\end{equation}
where $P_{ROF}(\lambda f,1/(\lambda\mu))$ is defined as the output of the ROF model applied to $\lambda f$ with a coefficient $1/(\lambda\mu)$ and is efficiently implemented by the Split Bregman iterations.
\end{proposition}
\begin{proof}
From Eq.~\ref{eq:gnorm} we have
\begin{eqnarray}
\partial J^*\left(\frac{v}{\mu}\right)-\lambda (f-v)\ni 0\\
\Leftrightarrow \lambda(f-v) \in \partial J^*\left(\frac{v}{\mu}\right)\\
\Leftrightarrow \partial J(\lambda(f-v))\ni \left(\frac{v}{\mu}\right)
\end{eqnarray}
Setting $w=\lambda(f-v)$ we get
\begin{eqnarray}
\partial J(w)\ni \frac{f-w/\lambda}{\mu}\\
\Leftrightarrow 0\in \partial J(w)-\frac{1}{\lambda\mu}(\lambda f-w)
\end{eqnarray}
This is equivalent that $w$ is a minimizer of $J(w)+\frac{1}{2\lambda\mu}\|\lambda f-w\|_2^2$ which is nothing less than the ROF model. 
Consequently, $\hat{w}=P_{ROF}(\lambda f,1/(\lambda\mu))$ and finally $\hat{v}=f-\frac{1}{\lambda}P_{ROF}(\lambda f,1/(\lambda\mu))$. \qed
\end{proof}

The new algorithm providing the cartoon + textures decomposition based on Eq.~\ref{eq:tvg} is presented in Algorithm.~\ref{alg:buv} where $P_{ROF}$ is computed by Algorithm.~\ref{algo:IROF}. The correspond MATLAB source code to compute $P_{ROF}$ is freely available in the Bregman Cookbook \cite{BregmanCookbook}.
\begin{algorithm}[!h]
\caption{Cartoon + textures decomposition algorithm based on Split Bregman iterations.}
\label{alg:buv}
\begin{algorithmic}
\STATE Initialisation: $u_0=v_0=0$
\WHILE {`` Not converged''}
\STATE Update $u$ by $u_{n+1}=P_{ROF}(f-v_n,\lambda)$
\STATE Update $v$ by $v_{n+1}=f-u_{n+1}-\frac{1}{\lambda}P_{ROF}(\lambda(f-u_{n+1}),1/(\lambda\mu))$
\ENDWHILE
\end{algorithmic}
\end{algorithm}

\section{Experiments}\label{sec:expe}
\begin{figure}[!t]
\includegraphics[width=0.99\columnwidth]{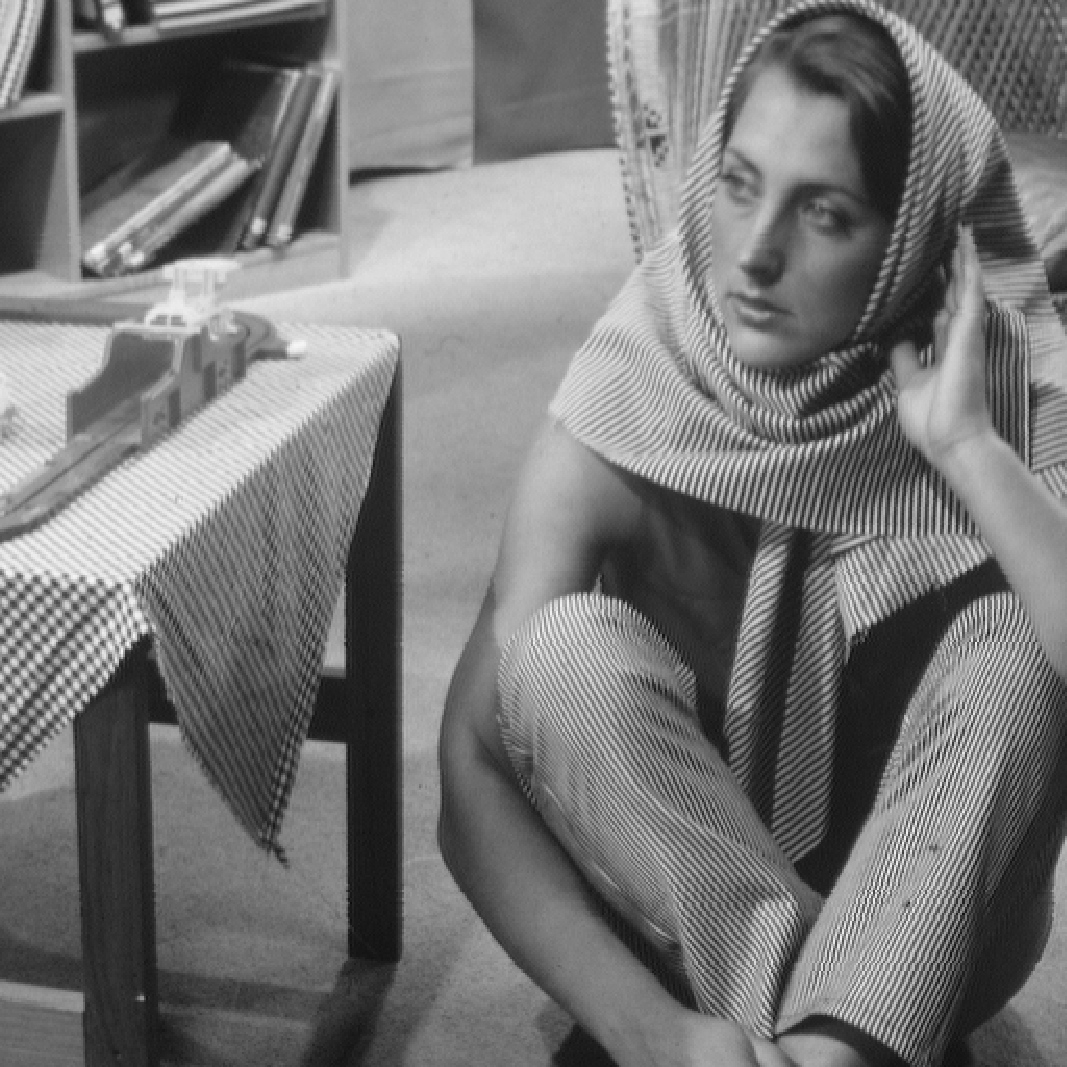}\\
\includegraphics[width=0.99\columnwidth]{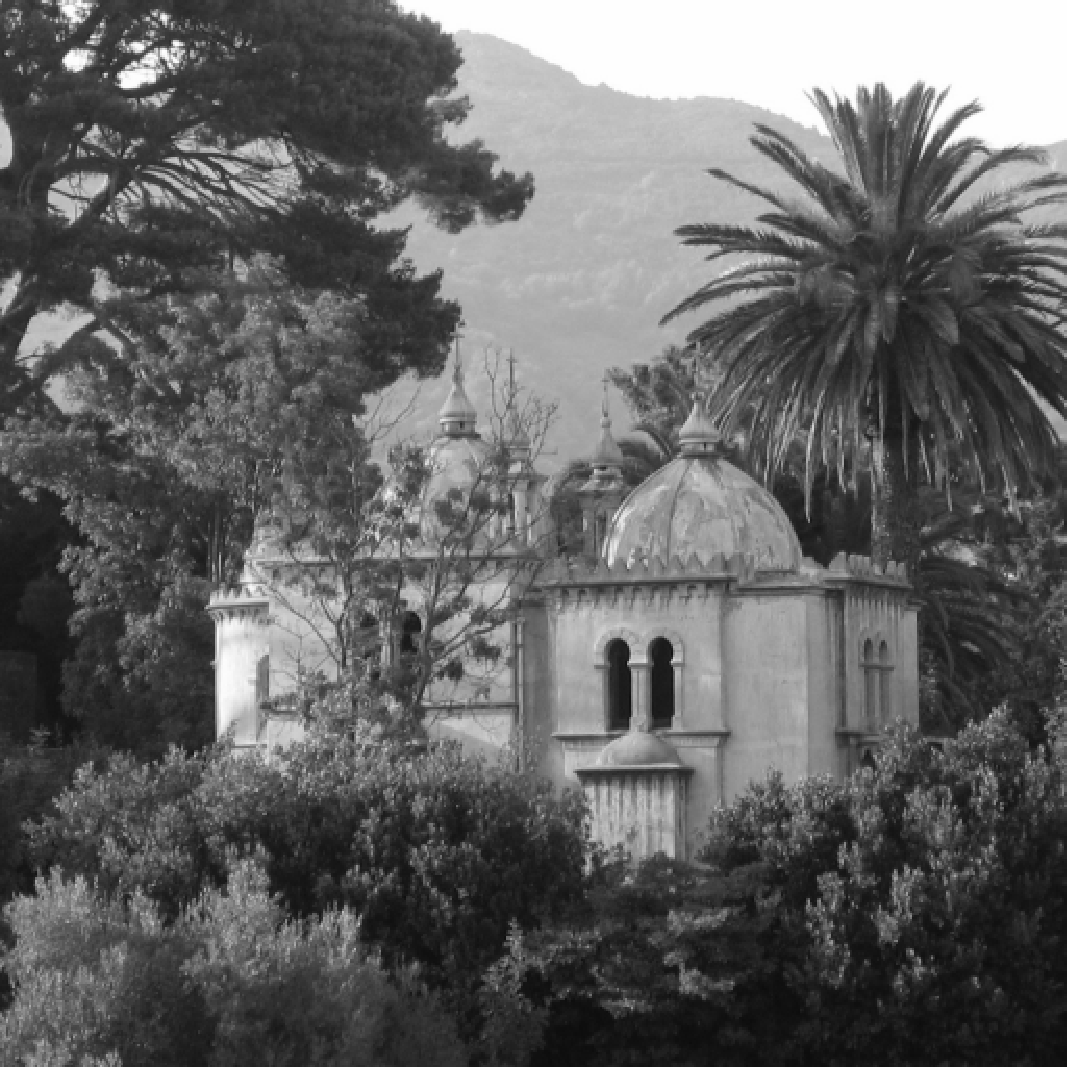}
\caption{Original images used as inputs of the decomposition algorithms.}
\label{fig:orig}
\end{figure}

\begin{figure}[!t]
\includegraphics[width=0.99\columnwidth]{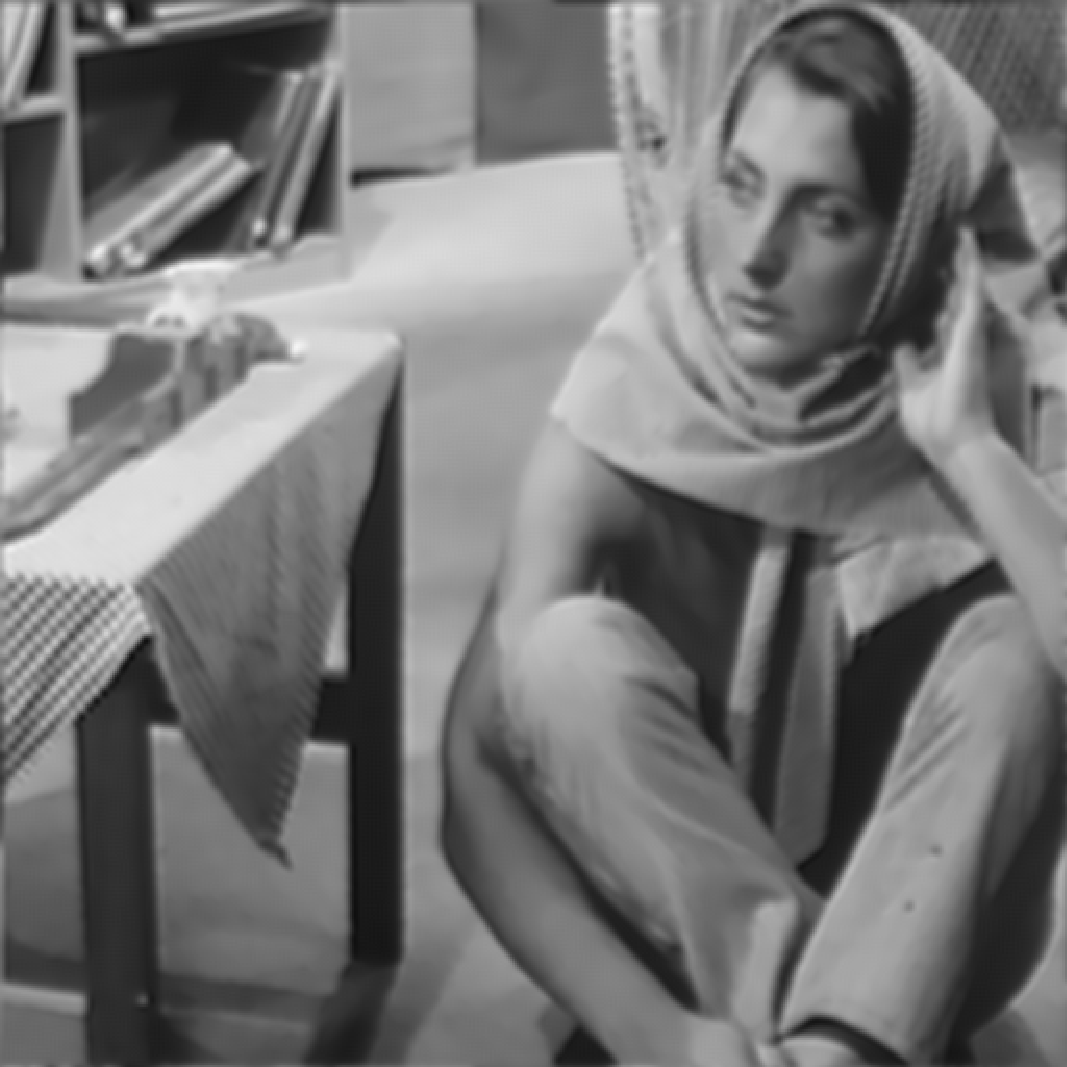}\\
\includegraphics[width=0.99\columnwidth]{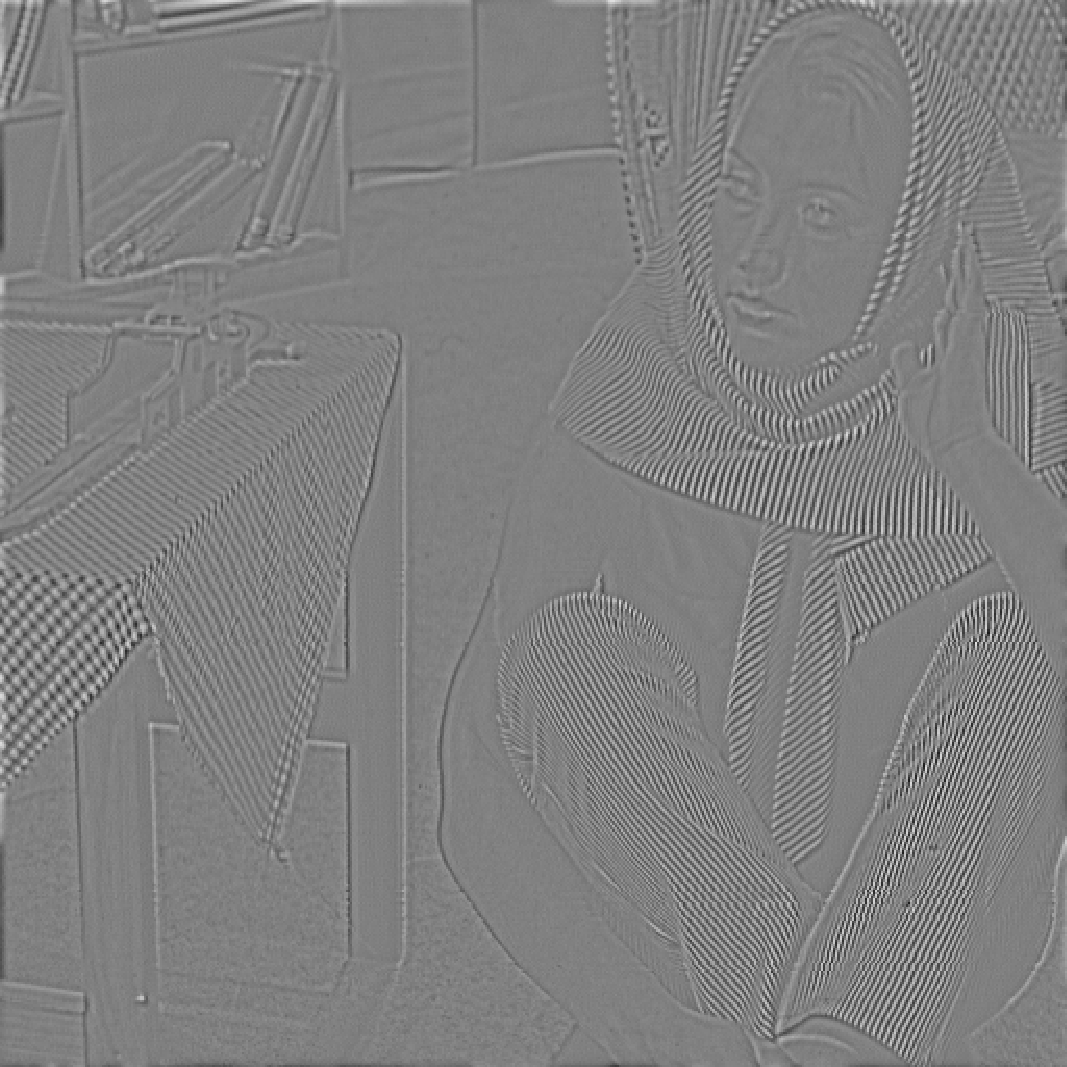}
\caption{Cartoon + textures parts obtained with $\lambda=\mu=1000$ for both images.}
\label{fig:dec1}
\end{figure}

\begin{figure}[!t]
\includegraphics[width=0.99\columnwidth]{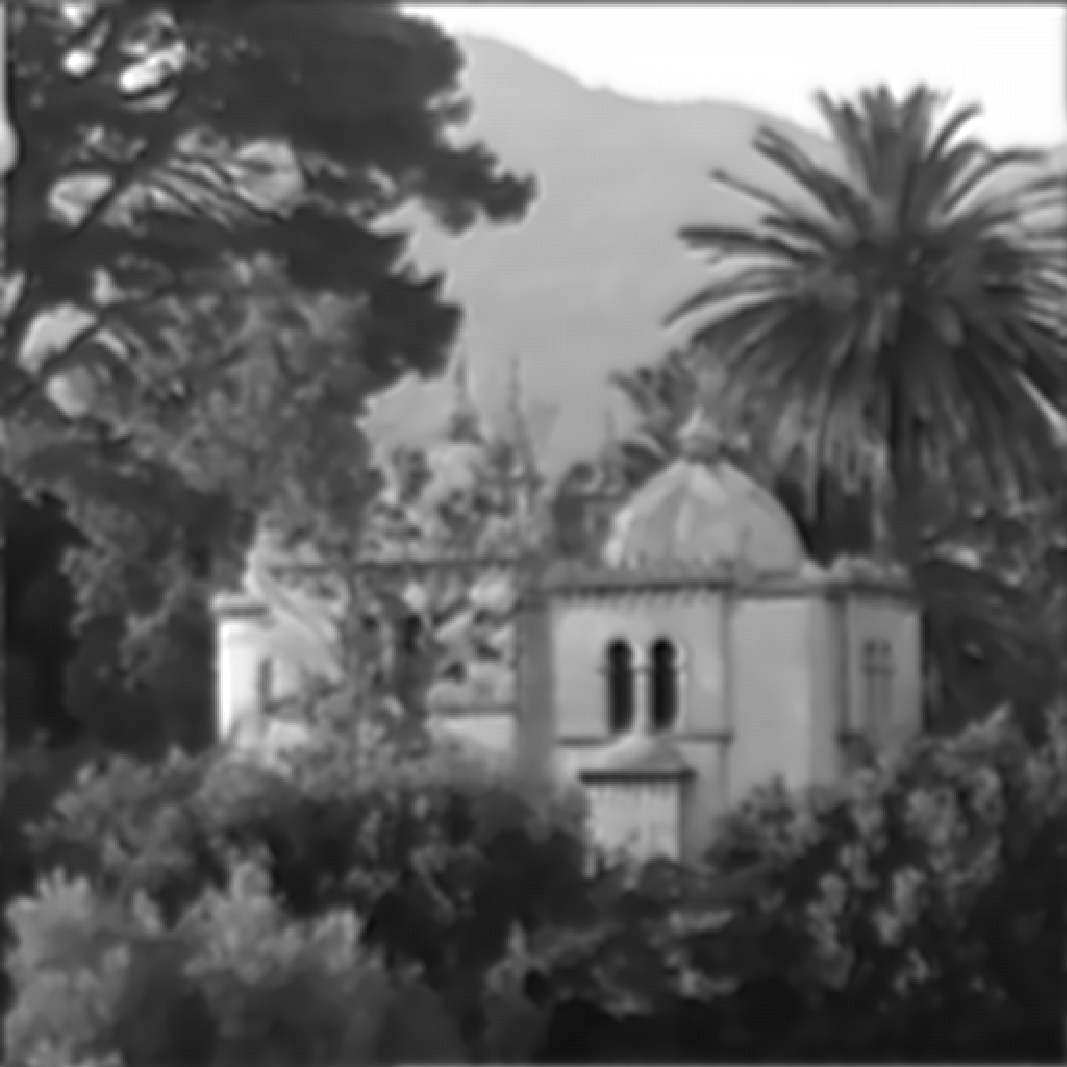}
\includegraphics[width=0.99\columnwidth]{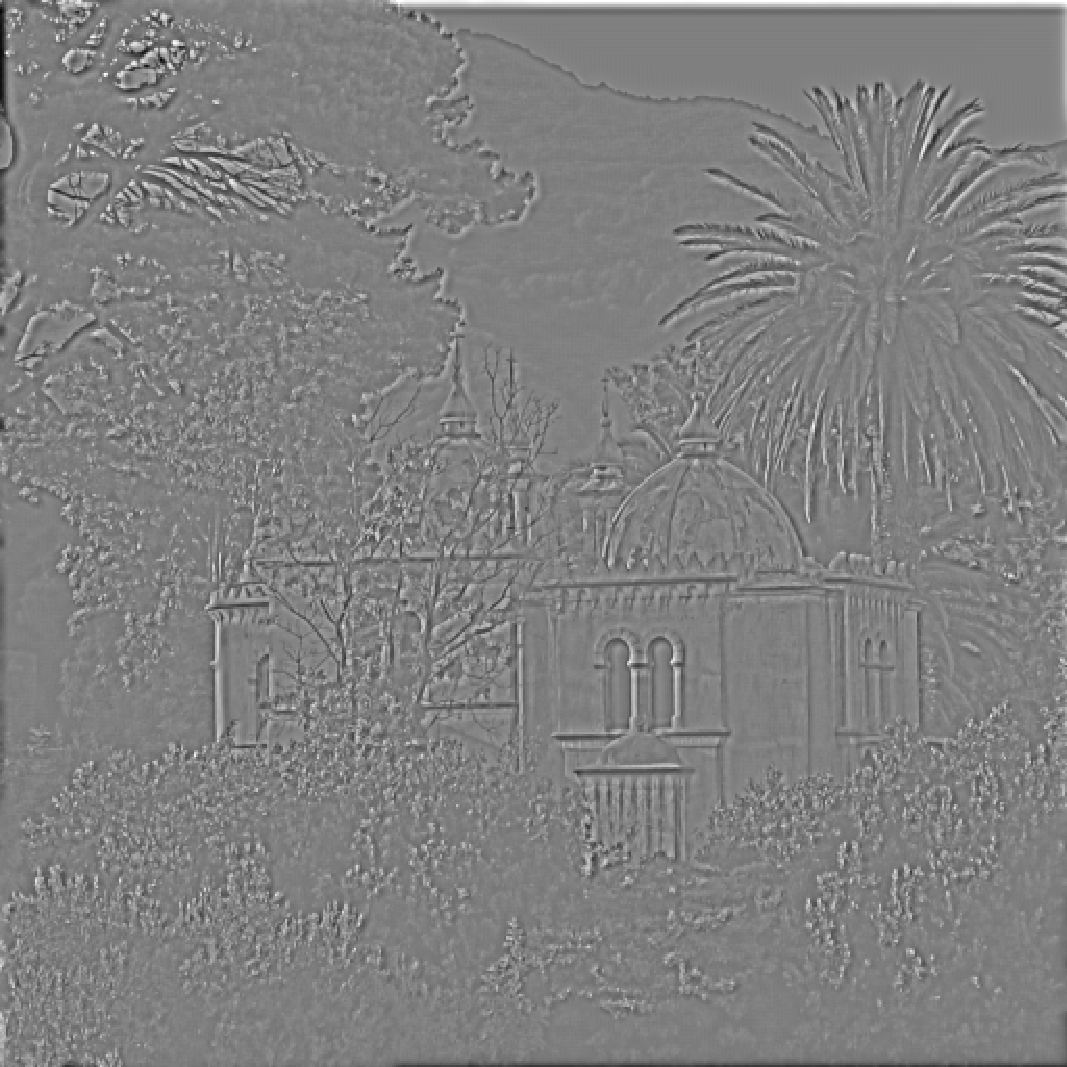}
\caption{Cartoon + textures parts obtained with $\lambda=\mu=1000$.}
\label{fig:dec2}
\end{figure}

\begin{figure}[!t]
\includegraphics[width=0.99\columnwidth]{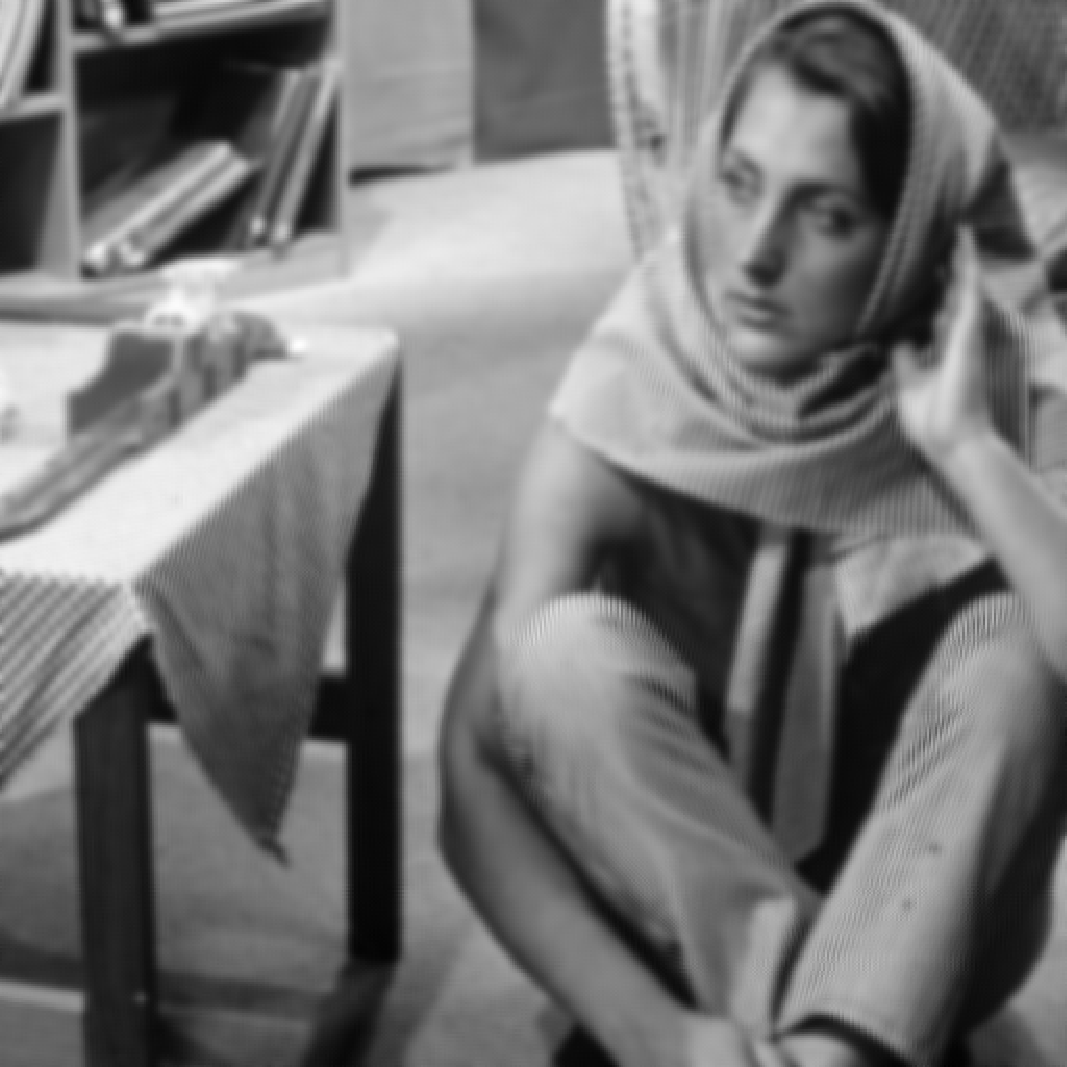}
\includegraphics[width=0.99\columnwidth]{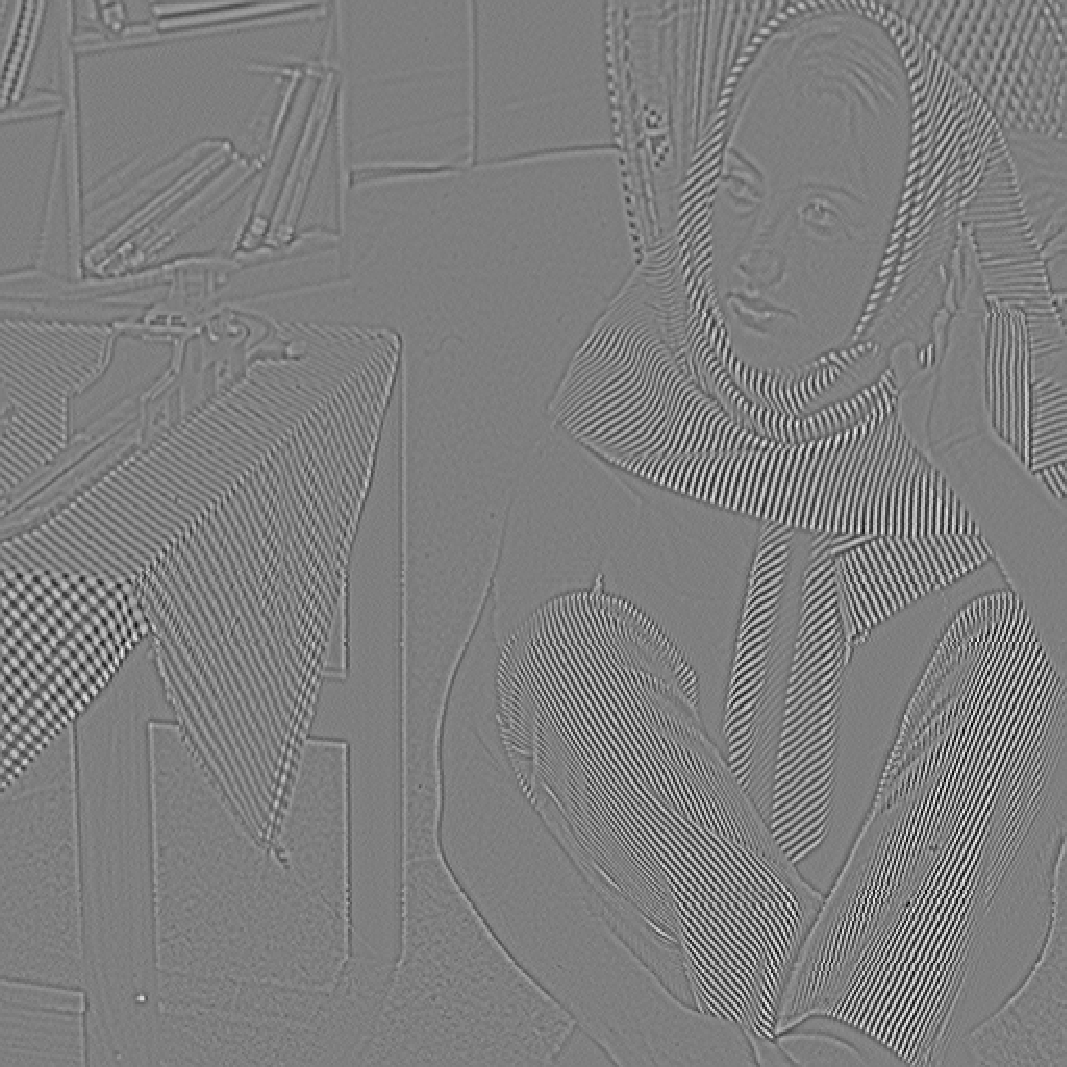}
\caption{Cartoon + textures parts obtained with $\lambda=\mu=1000$ from the nonlinear projectors.}
\label{fig:Pdec}
\end{figure}

In this section, we present the output of Algorithm.~\ref{alg:buv} applied on the two images depicted on Fig.~\ref{fig:orig}. In both experiments, we set $\lambda=\mu=1000$. Figures~\ref{fig:dec1} and \ref{fig:dec2} show the corresponding cartoon and textures parts obtained by the proposed algorithm. In order to compare with the results given by Aujol et al. algorithm, we show the output of the nonlinear projector method of Barbara on Fig.~\ref{fig:Pdec}. We see that the Bregman based algorithm performs well on the decomposition.\\
We ran many experiments to compare the speed of the two methods. These tests show that the number of iterations for both Chambolle's and Bregman's algorithms are, in average, the same. But Chambolle's projector needs more iterations to converge to $P_{G_{\mu}}$ than the Split Bregman needs to converge to $P_{ROF}$. Then globally, the new approach based on Bregman iterations is faster.

\section{Conclusion}\label{sec:conc}
In this paper, we propose an algorithm based on Split Bregman iterations to solve models based on Meyer's $G-$norm. The direct application is a dual approach to perform the cartoon + textures decomposition of an image. Experiments show the effectiveness of the algorithm and the use of Bregman iterations clearly improve the speed of the decomposition.

\begin{acknowledgements}
This work is supported by the following grants: NSF DMS-0914856, ONR N00014-08-1-119, ONR N00014-09-1-360.
\end{acknowledgements}

\bibliographystyle{IEEEbib}

\end{document}